\documentclass[sn-mathphys,Numbered]{sn-jnl}

\usepackage{graphicx}%
\usepackage{multirow}%
\usepackage{amsmath,amssymb,amsfonts}%
\usepackage{amsthm}%
\usepackage{mathrsfs}%
\usepackage[title]{appendix}%
\usepackage{xcolor}%
\usepackage{textcomp}%
\usepackage{manyfoot}%
\usepackage{booktabs}%
\usepackage{algorithm}%
\usepackage{algorithmicx}%
\usepackage{algpseudocode}%
\usepackage{listings}%
\usepackage{svg}
\usepackage{tikz}
\usepackage{subcaption}
\usepackage{tabularx}
\usepackage[sorting=none]{biblatex}
\bibliography{TreeCoderGraph}
 
\usepackage{makecell}
\usepackage{wrapfig}

\theoremstyle{thmstyleone}%
\theoremstyle{thmstyletwo}%

\theoremstyle{thmstylethree}%
\newtheorem{definition}{Definition}%

\begin{document}

\title[Article Title]{TreeCoders: Trees of Transformers}


\author{\fnm{Pierre} \sur{Colonna D'Istria}}\email{pierrecolodi@gmail.com}
\author{\fnm{Abdulrahman} \sur{Altahhan}}\email{a.altahhan@leeds.ac.uk}

\affil{\orgdiv{School of Computer Science, University of Leeds}}

\abstract{
In this paper, we introduce TreeCoders, a novel family of transformer trees. 
We moved away from traditional linear transformers to complete k-ary trees. Transformer blocks serve as nodes and generic classifiers learn to select the best child and route the sequence of tokens to a specific leaf. The selectors, moved outside the transformer blocks, allow for the use of a variety of architecture without further modifications. Furthermore, our proposed architecture supports sparse node activation due to the logarithmic complexity of a tree search. We validate our idea by testing a series of decoder-only tree transformers achieving competitive results across a diverse range of language datasets. Our study demonstrates that the proposed tree transformer model outperforms a size-equivalent linear transformer model 76\% of the time over a wide range of tree architectures. Furthermore, our proposed model naturally lends itself to distributed implementation.

}

\keywords{Transformers, Large Language Model, Tree, Language modeling}



\maketitle
\section{Introduction}\label{sec1}
Transformers have proved their effectiveness ever since their introduction \cite{Vaswani17}.
One development has been a rapid increase in the size of the models and the amount of data they consume, leading to models such as GPT-3 \cite{Gpt3} towering at 175 billion parameters. Those large and dense models impose heavy hardware requirements and longer inference time. Better management of those resources would be greatly beneficial. One possible solution to reduce inference time has been to promote sparsity in networks through the Mixture-of-Expert approach \cite{Jacobs1991AdaptiveMO}\cite{Jordan1993HierarchicalMO}. A tree is sparse by design, as shown in Figure \ref{fig:sparsity} and, therefore, an attractive candidate for scaling future models.
Large language models are increasingly expected to perform on a variety of tasks and datasets. Textual data varies greatly in terms of structure, intent, content, formality, language, and organization. Nevertheless, the underlying language apparatus of these contents can be hierarchically exploited based on their common features.
This is our second motivation to investigate moving from a linear structure to a tree-like one. The final motivation is the inherently explainable nature of decision trees. Bringing about explainability to large language models would be a formidable step forward in itself but should also help with fine-tuning, safety, and alignment.  

\begin{figure} 
  \centering
  \begin{subfigure}{0.6\textwidth}
  \begin{minipage}{0.48\textwidth}
\caption{A sequence going through one path in the tree, from root decoder A to leaf decoder K.}
  \begin{tikzpicture}[level distance=1.5cm, level 1/.style={sibling distance=4cm}, level 2/.style={sibling distance=2cm}, level 3/.style={sibling distance=1cm}]
    \node[circle, draw] (root) {A}
      child {
        node[circle, draw] {B}
        child {
		node[circle, draw, dashed] {D}
		edge from parent [dashed]
          child {
            node[circle, draw, dashed] {H}
          }
          child {
            node[circle, draw, dashed] {I}
          }
        }
        child {
		node[circle, draw] {E}
		child {
          	node[circle, draw, dashed] {J}
		edge from parent [dashed]
		}
		child {
 		node[circle, draw] {K}
 		}
        }
      }
      child {
	node[circle, draw, dashed] {C}
	edge from parent [dashed]
        child {
		node[circle, draw] {F}
		child {
          	node[circle, draw, dashed] {K}
		edge from parent [dashed]
		}
		child {
 		node[circle, draw] {L}
 		}
        }
        child {
          node[circle, draw] {G}
		child {
          	node[circle, draw, dashed] {M}
		edge from parent [dashed]
		}
		child {
 		node[circle, draw] {N}
 		}
        }
      };
  \end{tikzpicture}
  \end{minipage}\hfill
  \begin{minipage}{0.48\textwidth}
  \end{minipage}\hfill
\end{subfigure}
\par\bigskip

\begin{subfigure}{\textwidth}
  \begin{minipage}{0.48\textwidth}
  \caption{Total number of nodes in the tree. \smallskip    }
  \label{tab:sparse-activation}
\medskip
  \begin{tabularx}{\textwidth}{c @{\hspace{1cm}} *{4}{X} @{}}
    \toprule
    Height & \multicolumn{4}{c}{\thead{Branching Factor}} \\
    \cmidrule(){2-5}
    & 1 & 2 & 3 & 4 \\
    \midrule
    1	& 2	&3	&4	   & 5\\
    2	& 3	&7	&13	   & 21\\
    3	& 4	&15	&40	   & 85\\
    4	& 5	&31	&121   & 341\\
    5	& 6	&63	&364   & 1365\\
    \bottomrule
  \end{tabularx}
\end{minipage} 
\hfill 
\begin{minipage}{0.48\textwidth}
  \caption{Percentage of parameters used by a token.}
  \label{tab:sparse-activation1}
  \begin{tabularx}{\textwidth}{c @{\hspace{1cm}} *{4}{X} @{}}
    \toprule
    Height & \multicolumn{4}{c}{\thead{Branching Factor}} \\
    \cmidrule(){2-5}
    & 1 & 2 & 3 & 4 \\
    \midrule
    1 & 100.0 & 66.7 & 50.0 & 40.0 \\
    2 & 100.0 & 42.9 & 23.1 & 14.3 \\
    3 & 100.0 & 26.7 & 10.0 & 4.7 \\
    4 & 100.0 & 16.1 & 4.1 & 1.5 \\
    5 & 100.0 & 9.5 & 1.6 & 0.4 \\
    \bottomrule
  \end{tabularx}
\end{minipage}
\end{subfigure}
\caption{A tree structure allows for a sparse activation of the network. The sparsity will also grow with the tree }
\label{fig:sparsity}
\end{figure}

\section{Related Work}
%


The history of the MOE starts with the seminal paper Adaptive Mixtures of Local Experts\cite{Jacobs1991AdaptiveMO} where they introduced the idea of an expert network and of a gating network. Both were feedforward neural networks. The gating network is in charge of assigning weights to the experts' outputs, which leads to experts handling different tasks or parts of the input space .
Another important early paper is "Hierarchical mixtures of experts and the EM algorithm" by Jordan and Jacobs\cite{Jordan1993HierarchicalMO}.
In their work, the results of two sets of experts and gating networks are then gated again. This creates a two-level tree-structure network.
Our models are also hierarchical and tree-like but we flip the structure upside down, starting with a single expert and progressively branching out to multiple experts. It allows us to sequentially refine the subsets of tasks handled by experts.
Conditional computation, as explored by Bengio et al. \cite{BengioBPP15}\cite{BengioLC13} or  Davis and Arel \cite{davis2013low} among others, is the idea of conditioning the operations performed by the model on specific inputs. The goal of such an approach is to reduce resource consumption and increase computation speed.
Mixture-of-Experts and Conditional Computation eventually found their way to the field of NLP, first using massive LSTM \cite{ShazeerMMDLHD17} and then introduced to Transformers witht the GShard model \cite{Gshard}. This model uses sparse MOE and automatic sharding to reach 600 billion parameters. Another notable model is GLaM \cite{glam} with a parameter count of 1.2 trillion.
The Switch Transformer \cite{Fedus2021} explores the idea of sending the tokens to only one expert, instead of the traditional top-k.
It comforted us in our choice of sending inputs to individual nodes.  
The interest in Mixture-of-Experts persists, with the introduction of models such as NLLB-200 \cite{costa2022no}, or Mixtral 8x7B \cite{jiang2024mixtral}.

Besides language mixture-of-Experts have been applied to the speech recognition \cite{speechmoe1} \cite{speechmoe2}, to image classification  \cite{lou2022crosstoken}, to ViT architecture \cite{VitMoe}, or even multimodal learning \cite{mustafa2022multimodal}.
For an in-depth review Mixture of Experts, see Fedus et al. \cite{fedus2022review} .

\section{Definitions}

\begin{definition}[Node]
A node is a transformer block composed of one or more layers.  
\end{definition}
The number of layers doesn't change the node count. \\

\begin{definition}[Height]
The height of a TreeCoder is the number of nodes (i.e transformer blocks) from root to leaf. We denote it later on with the letter h. 
\end{definition}
The number of encoder or decoder layers does not influence a TreeCoder height.\\

\begin{definition}[Number of layers]
The number of layers refers to the number of layers per node. We denote it by dec later on when speaking of decoder layers. 
\end{definition}
A tree architecture can be partly defined with the tuple (h, dec).\\

\begin{definition}[Token Path Length]
The token path length is the number of transformer layers a token goes through from root to leaf, i.e. from input to output. 
\end{definition}
Two trees of different heights can have the same token path length.
The path length is an important measure as it allows us to make comparisons with linear transformers but also group trees and observe trends or the influence of height. \\

\begin{figure}[H]
\centering
\includegraphics[width=250pt]{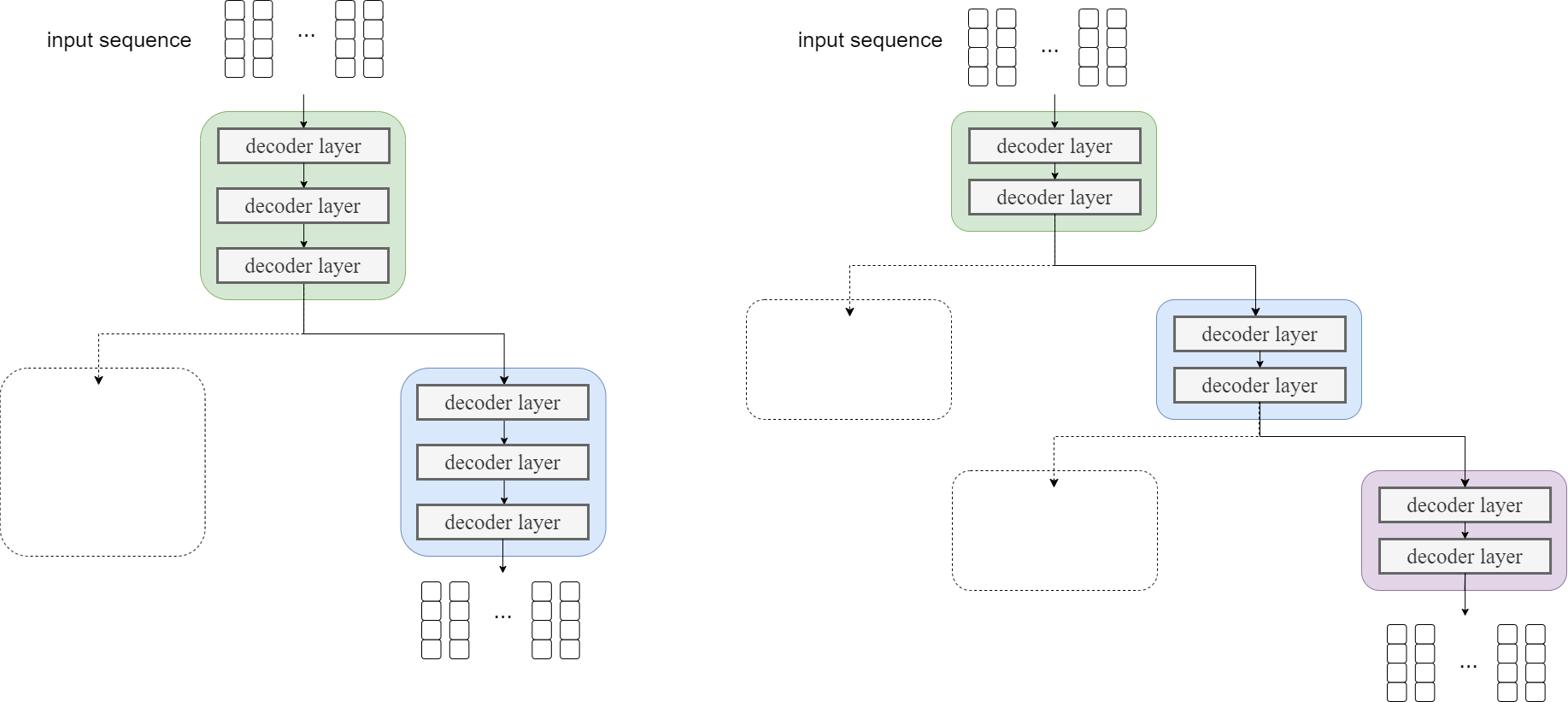}
\caption{Example of two different tree architectures where the tokens go through the same number of decoder layers. On the left is a tree (h, dec) = (1, 3) of height  1, with 3 decoder layers per node. On the right is a tree of height 2 with 2 decoder layers per node. In both cases, the path length is 6. }
\label{fig:path_length}
\end{figure}

\section{The Transformer Tree Model }
A TreeCoder is composed of two main components: the transformer blocks and the selectors. 
We build a k-ary tree where each node is a stack of one or more decoder (or encoder) layers.
The root node receives the same input as a classic transformer and performs the same operations. 
The result is used as input for the first branch selector. They will have as many outputs as the prespecified tree branching factor (e.g 2 for a binary tree, 3 for a ternary tree, etc).
Once a child has been selected, the root node's output is routed to it. The process repeats until a leaf node is reached. 
The leaf node will perform the usual transformer-block operations and will finally output probabilities. That output will be used to compute the loss and perplexity.

\begin{wrapfigure}{r}{0.5\textwidth}
    \begin{center}
        \includegraphics[width=0.4\textwidth]{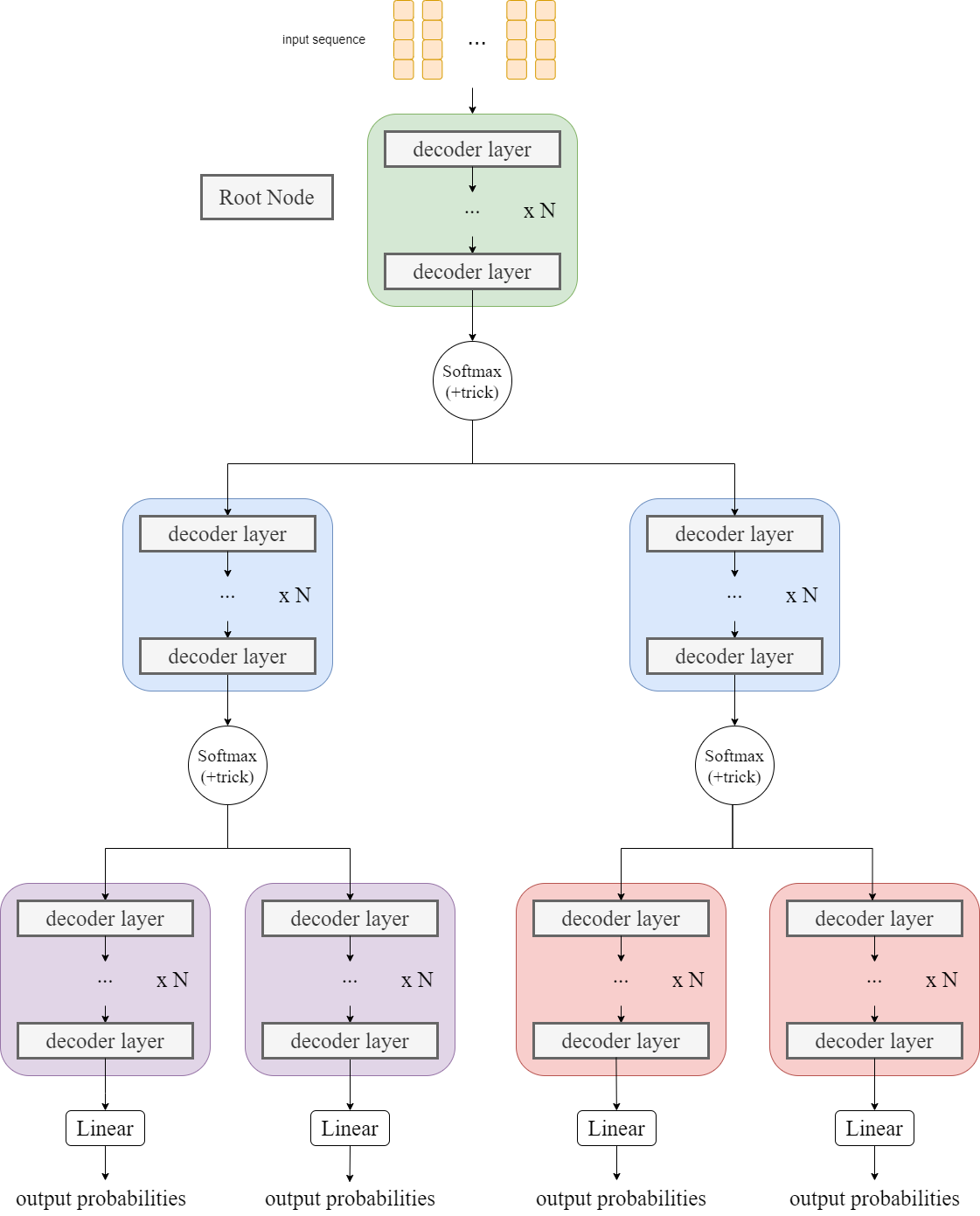}
        \caption{Example of a binary tree of height 2 and N decoder layer per node.}
        \label{fig:tree-example}
    \end{center}
\end{wrapfigure}

Figure \ref{fig:tree-example} illustrates an example of a binary TreeCoder of height 2 with N decoder layers per node followed by a softmax selector. We take inspiration from Switch Transformer \cite{Fedus2021}, where tokens are sent to the top-1 expert and forward values to one child only. Where other Mixture-of-expert models regroup, we send intermediate results to another set of independent experts. Importantly, our proposed model differs from their model in that we have a diverse structure with an ever-growing number of experts as we go down the tree. 

This framework can support encoder-only models like BERT \cite{Bert}, decoder-only like the GPT \cite{gpt1}\cite{gpt2}\cite{Gpt3}  series of models, or a mix of both. One can imagine combining pre-trained models as nodes as long as the different formats are compatible. The routing decision can be made by something as simple as a binary classifier or by more advanced techniques such as Monte-Carlo Tree Search (MCTS) or reinforcement learning.

We will focus our experiments on decoder-only models due to the success of such models in both scientific literature and production, with varying heights and number of layers per block. Finally, we will employ a softmax classifier as a decision-maker due to its simplicity and differentiability with an extra programming trick to allow for the built-in automatic differentiation to work with our proposed architecture.

\begin{table}[!ht]
    \caption{Total number of parameters (in millions) used by trees of different heights and number of decoder layers}
    \label{tab:selec-percent1}

    \begin{tabularx}{\textwidth}{l *{6}{X} }
    \toprule
	 & & \multicolumn{5}{c}{\thead{Height Of The Tree}} \\
    \cmidrule(){3-7}
    Decoder& &  1       & 2       &     3    & 4  	     &    5        \\
    \midrule
    1 && 71		& 154	& 322	& 656	& 1325 \\
    2 && 109	& 243	& 511	& 1046	&      \\
    3 && 146	& 331	& 699	& 1437	&      \\
    4 && 184	& 419	& 888	&   	&      \\
    5 && 222	& 507	& 1077	&   	&      \\
    6 && 260	& 595	&   	&   	&      \\
    7 && 297	&    	&   	&   	&       \\
    8 && 335	&   	&   	&   	&      \\
    \bottomrule
    \end{tabularx}
\end{table}
Our largest model is only 1.4 billion parameters, which is small by today's standard, smaller than GPT2 1.5 billion or LLaMA's smallest model with 7 billion parameters \cite{llama1} \cite{llama2}. More importantly, out of those 1.4 billion, we only use 16.1\% at inference, as seen in Figure {tab:sparse-activation} 
\subsection{Possible Architectures }
In Figure \ref{fig:topo}, we illustrate the multiple architectures that are possible with this idea. We can build an encoder-only tree, like in Figure \ref{fig:topo}\subref{fig:subfig-a}, or decoder-only as in Figure \ref{fig:topo}\subref{fig:subfig-b}.
We can modify the original transformer architecture \cite{Vaswani17}  by attaching a tree of decoders to the encoder as in Figure \ref{fig:topo}\subref{fig:subfig-c}. Inversely, we could build a tree of encoders and add linear decoders to each leaf. Finally, we could combine both and have a tree of encoders with a tree of decoders attached to each leaf as in Figure \ref{fig:topo}\subref{fig:subfig-e}.

\begin{figure}
\caption{Some of the different topologies possible with our method}
\label{fig:topo}
  \begin{subfigure}{1\textwidth}
  \begin{minipage}{0.33\textwidth}
	\caption{Encoder only}
	\label{fig:subfig-a}
	\medskip
	\begin{tikzpicture}[level distance=1cm ]
	  \node {Encoder}
	    child { node {Selector}
	      child { node {Encoder} }
	      child { node {Encoder} }
	    };
	\end{tikzpicture}
  \end{minipage}
  \begin{minipage}{0.33\textwidth}
	\caption{Decoder only}
	\label{fig:subfig-b}
	\medskip
	\begin{tikzpicture}[level distance=1cm ]
	  \node {Decoder}
	    child { node {Selector}
	      child { node {Decoder} }
	      child { node {Decoder} }
	    };
	\end{tikzpicture}

  \end{minipage}\hfill
\begin{minipage}{0.33\textwidth}

  	\caption{Encoder followed by tree of decoders}
	\label{fig:subfig-c}
	\medskip
	\begin{tikzpicture}[level distance=1cm ]
	  \node {Encoder}
	    child { node {Selector}
	      child { node {Decoder} }
	      child { node {Decoder} }
	    };
	\end{tikzpicture}

\end{minipage} 

\end{subfigure}

\bigskip\bigskip\bigskip

\begin{subfigure}{\textwidth}
  
\begin{minipage}{0.47\textwidth}
	
	\begin{tikzpicture}[level distance=1cm ]
		
	  \node {Encoder}
	    child { node {Selector}
		child { 
			node {Encoder} 
				child { node {Decoder} }
		}
	      child { 
			node {Encoder}
				child { node {Decoder} }
		}
	    };
	\end{tikzpicture}
	\bigskip\bigskip\bigskip\bigskip\bigskip\bigskip
	\caption{Tree of encoders followed by linear decoders}
	\label{fig:subfig-d}
	
\end{minipage}
\hfill
\begin{minipage}{0.45\textwidth}
	
	\begin{tikzpicture}[level distance=1cm, level 1/.style={sibling distance=3cm},  level 5/.style={sibling distance=1.5cm}]
	 
	  \node {Encoder}
	    child { node {Selector}
		child { 
			node {Encoder} 
			child { 
				node {Decoder}
		    			child { 
						node {Selector}
		      				child { node {Decoder} }
		      				child { node {Decoder} }
		    			}
			}
		}
	      child {
			node {Encoder}
			child { 
				node {Decoder}
		    			child { 
						node {Selector}
		      				child { node {Decoder} }
		      				child { node {Decoder} }
		    			}
			}
		}
	    };
	\end{tikzpicture}
	\medskip
	\caption{Tree of encoders  followed  by tree of decoders}
	\label{fig:subfig-e}
	\end{minipage}
\end{subfigure}
\end{figure}

\subsection{Selectors}
Selectors are fully connected neural networks ending with a softmax. 
It takes as input a sequence of tokens of shape $context\ length * embedding\ size$. We then perform mean-pooling.
The selector architecture can vary to suit one's needs, but the smallest model tested is a fully connected neural network with a single hidden layer whose size is eight times the input size.
SwiGLU \cite{shazeer2020glu} is used as an activation function.
We used a softmax to accommodate binary as well k-ary trees.

The index of the logit with the highest value determines which child will receive its parent's output as input.
We also return the logit divided by the logit value. Although always equal to one, it retains in PyTorch the information necessary to perform backpropagation later. 

\begin{equation}
grad\_trick = 
    \frac{max\_probality}{max\_probality.detach()} 
\end{equation}

Fedus et al. \cite{Fedus2021} refer to this top-1 routing as Switch Routing.
A significant difference between our method and the Switch Transformer is that we send complete sequences instead of a single token to the expert child, further reducing the routing complexity.
Differentiability is maintained by multiplying the transformer block output by the logit value divided by itself. In our experiments, it improves performance over multiplying by the logits.

        

\begin{figure}[!ht]
\centering
\includegraphics[width=\textwidth]{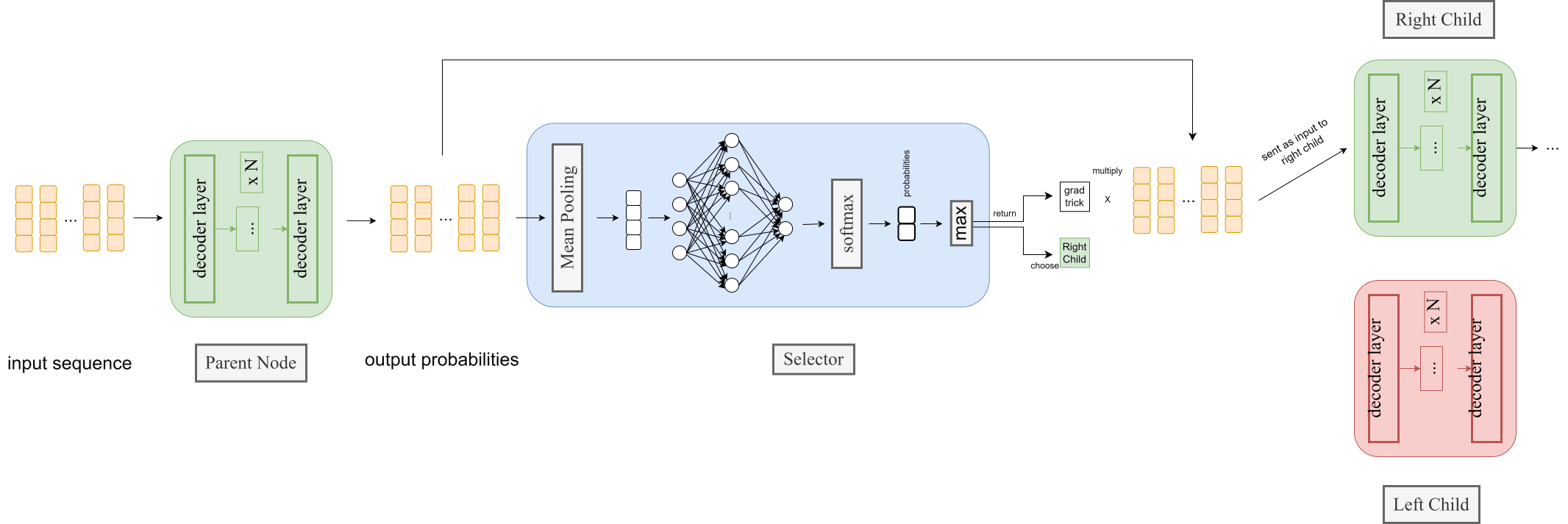}
\caption{Path of a sequence of tokens through a transformer node, a selector, and then sent to the selected child node.}
\end{figure}

\begin{algorithm} 
      \caption{Traversing the tree of a TreeCoder}\label{algo1}
      \hspace*{0pt} \textbf{Input} Sequence of tokens \\
      \hspace*{0pt} \textbf{Output} Output probabilities \\
      \hspace*{0pt} \textbf{Parameters} \\
            \hspace*{6em} height \Comment{Height of tree} \\
            \hspace*{6em} Decoders  \Comment{Tree of decoders}\\
            \hspace*{6em} Selectors \Comment{Tree of selectors}

      \begin{algorithmic}[1]
        \State $level \Leftarrow 0$ 
        \State $node \Leftarrow root$  
        
        
        \While{ $level <   height $}
        \State $output\Leftarrow Decoders[node](input)$
        \State $selected\_child,grad\_trick\Leftarrow Selectors[node](output)$
        \State $node \Leftarrow selected\_child$
        \State $input \Leftarrow output \times grad\_trick$
        \State $level \Leftarrow level + 1$
        \EndWhile
        
      \end{algorithmic}
    \end{algorithm}

\subsection{The Transformer-Selector Nodes}

The nodes are the usual decoders or encoders and would be referred to as experts in a Mixture-of-Experts context. In this paper, we focus on decoder-only models.
Our goal was not to develop a new decoder architecture and therefore we based our implementation on already successful models. Namely, we started off on a GPT-like decoder-only architecture. 
Given the results of the recently released LLama2 \cite{llama2}, we decided to apply some of the design choices. More specifically, we apply pre-normalization with RMSNorm  \cite{zhang2019root} and the SwiGLU activation function \cite{shazeer2020glu}. We did not use grouped-query attention \cite{ainslie2023gqa} and rotary positional embeddings \cite{su2023roformer}. We use an embedding size of 1024 and a context length of 128.

\subsection{Hyperparameters Settings}

\textbf{Tokenizer} We process our data with the byte pair encoding tokenizer from SentencePiece \cite{kudo2018sentencepiece}. The vocabulary contains 8k tokens.
We used a character coverage of 1, the identity normalization rule, split digits, allowed whitespace-only pieces, and byte fallback. \\\\
\textbf{Learning Rate} The base learning rate is 3e-4. We start with a 2,000-step warmup followed by a cosine annealing with a warm restart scheduler.  \\\\
\textbf{Dropout} We apply a fixed dropout rate of 0.1, gradient clipping of 1, and a weight decay of 10\%. \\\\
\textbf{Optimizer} We use the AdamW optimizer \cite{adamw} with a $\beta_1$ of 0.9 and a $\beta_2$ of 0.95. \\\\
\textbf{Data} Training happened on several datasets, namely Wikitext2, Wikitext-103 \cite{wikitext}, PenTreeBank \cite{ptb}. All training occurred on one dataset at a time unless mentioned otherwise. We imported them using the torchtext.datasets module.
We added beginning- and end-of-sequence tokens and padding.
The split for the training, validation, and test sets is the one provided by the module.
We used a batch size of 16. \\\\
\textbf{Seed} The seed was chosen randomly but fixed for the sake of fairness in the comparison of results.\\\\
\textbf{Training}
All our models are trained from scratch. All were trained on a single GPU, an A100 on a cluster.  We ran for 20 epochs, saving the model if it improved upon validation perplexity. The minimum was usually found around the fifteen epoch mark. We computed the test perplexity based on the parameters of the last saved model. 

\section{Difference with Mixture of Experts}
We will try to illustrate some of the major differences with Mixture of Experts.
For instance in the GLaM model \cite{glam},  "each input token is dynamically routed to two selected expert networks out of 64" (Andrew M Dai and Nan Du, 2021).
In our method, the gating occurs outside of the classic layers and not inside. We do not replace the feedforward network with experts (although we could).


\begin{figure}[!ht]
\centering
\includegraphics[width=250pt]{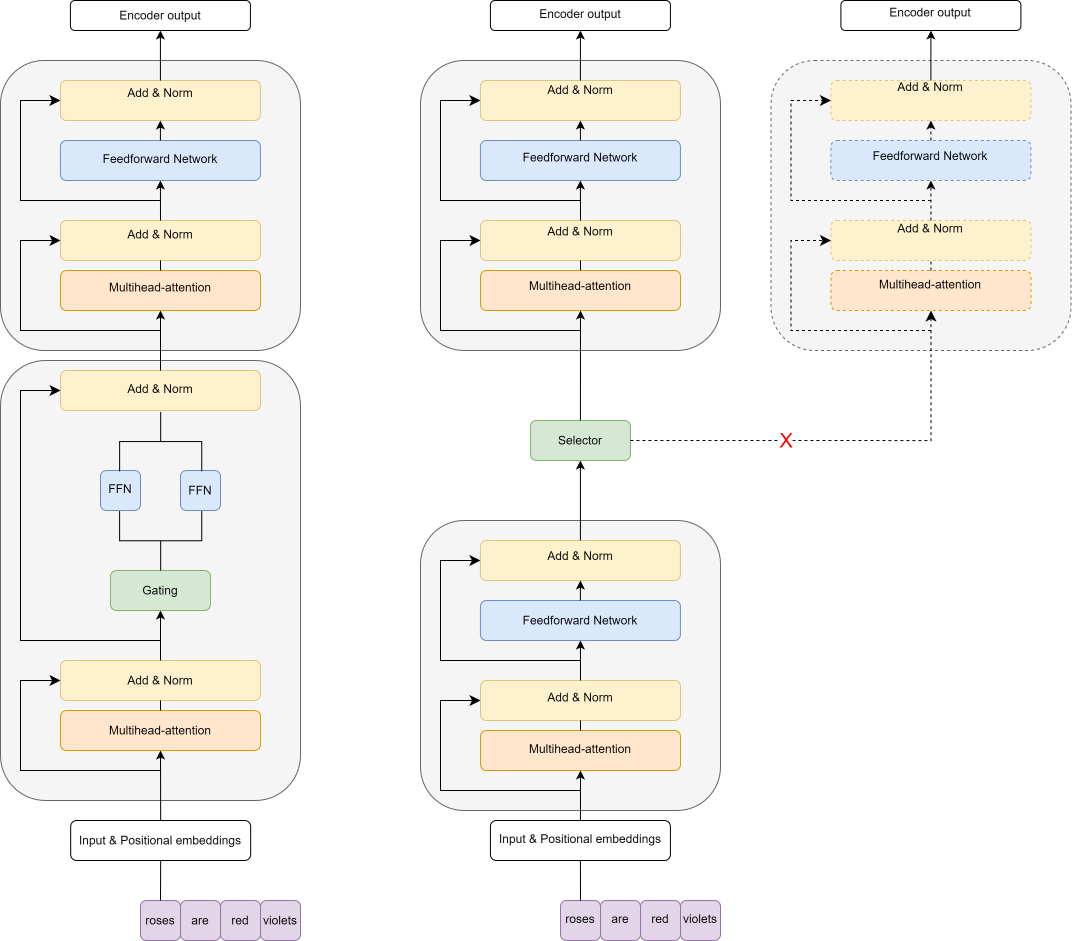}
\caption{One difference between Mixture-of-Expert and a TreeCoder. GlaM on the left, and on the right, our new architecture. Notice the gating happens outside of the layer.}
\label{fig:moe_v_tree}
\end{figure}

Also, while the "final learned representation of a token will be the weighted combination of the outputs from the two experts" for GLaM \cite{glamblog}, our representation will exclusively be the result of a single path. Once a path is taken (i.e an expert is chosen) a subset of others is de facto excluded and unreachable.
Another major difference is how tokens are dispatched. In the current literature,  single tokens are gated to the experts while we send a whole sequence to the selectors.
\section{Experimental Results}
While achieving high performance is undoubtedly a primary objective, our investigation goes beyond it.
Hence, we investigate our TreeCoders performance across diverse architectural configurations by systematically varying the tree height, the number of decoders per node,  the selector size and the branching factor.
The baseline is to identify configurations where the proposed model exhibits comparable or better capabilities than a standard linear transformer. More desirably, we want to show that the tree-based architecture consistently outperforms a transformer of equivalent size, thus mitigating the potential confounding factor of model complexity contributing to performance gains. Identifying configurations that support performance improvements strengthens the theoretical and practical relevance of our model. 
Finally, to ensure the selectors are actually learning a suitable routing behavior, i.e. sending down outputs to the better subtree, we compare our softmax selector with a random selector that chooses a child at random. We also study the selector's behavior with respect to its size. We expect a bigger model to improve on the final test perplexity. The question is of import as the selector can account for a significant amount of parameters.

Across 57 experiments, we report that a tree outperforms its linear counterpart 64.9\% of the time. Also, 76.2\% of the time we can find at least one tree with a given number of layers outperforming the linear transformer. Therefore, if one were to choose an architecture at random, one would see an improvement in perplexity 64.9\% of the time, with all the possible improvements a tree would bring. In the same way, if one were to test the few equivalent architectures and chose the best one, it would be an improvement 76.2\% of the time. 

\begin{table}[]
    \caption{Test perplexity of our models on different datasets.}
    \label{tab:perf_summary}
    \centering
    \begin{tabular}{llll}
    \toprule
                 & Test Perplexity &  Architecture (h, \#dec) & Parameter Count \\ \hline
    Wikitext2    & 30.23           & (1, 5)                                                                   & 222M            \\
    PennTreeBank & 37.56           & (1, 4)                                                                   & 184M            \\
    WT2+PTB      & 29.52           & (1, 7)                                                                   & 297M            \\  
    Wikitext103  & 15.79           & (3, 3)                                                                   & 699M            \\ \hline
    \end{tabular}
\end{table}

\subsection{Performance for different heights and decoder layers }
Table \ref{tab:ppl-h-dec} shows the performance of our model for different tree heights and decoder layers per node.
As expected, adding decoder layers for a given tree height (reading the table vertically) improves the perplexity up to a certain threshold, beyond which the performance improvement stops.

\begin{figure} 
\begin{center}
     \subfloat[WikiText2] { 
        \subfloat[Test perplexity]{
            \includegraphics[width=0.48\textwidth]{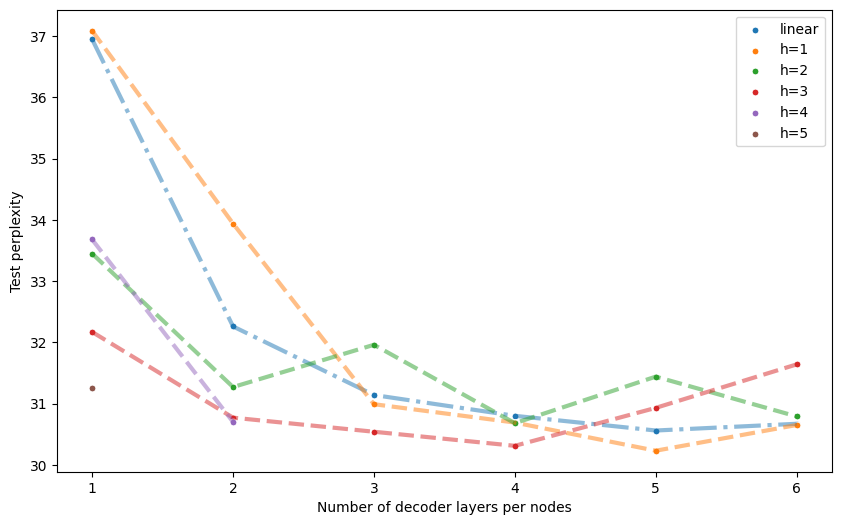}
            \label{fig:perf_height_combined}
        }
        \subfloat[Linear transformer vs best tree]{ 
            \includegraphics[width=0.48\textwidth]{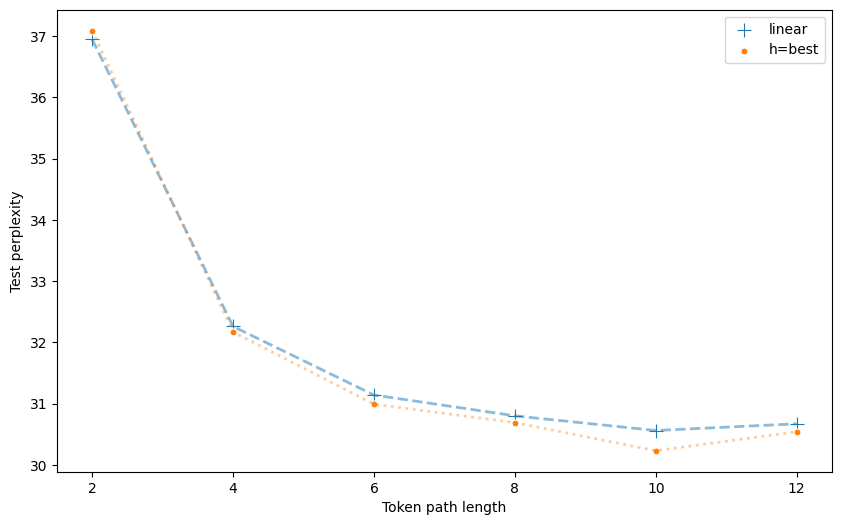}
            \label{fig:perf_linear_vs_best_h_combined}
        }
    }
    
    \subfloat [PennTreeBank]{ 
        \subfloat[Test perplexity]{
            \includegraphics[width=0.48\textwidth]{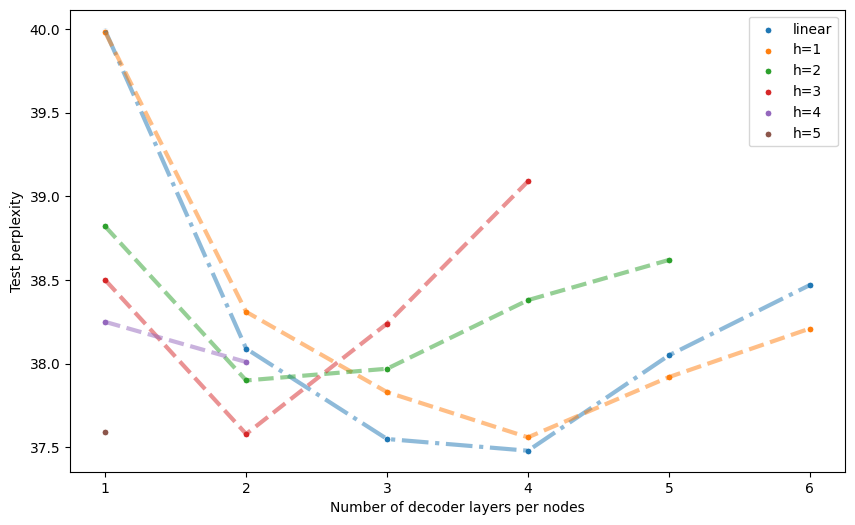}
            \label{fig:perf_height_combined}
        }
        \subfloat[Linear transformer vs best tree]{ 
            \includegraphics[width=0.48\textwidth]{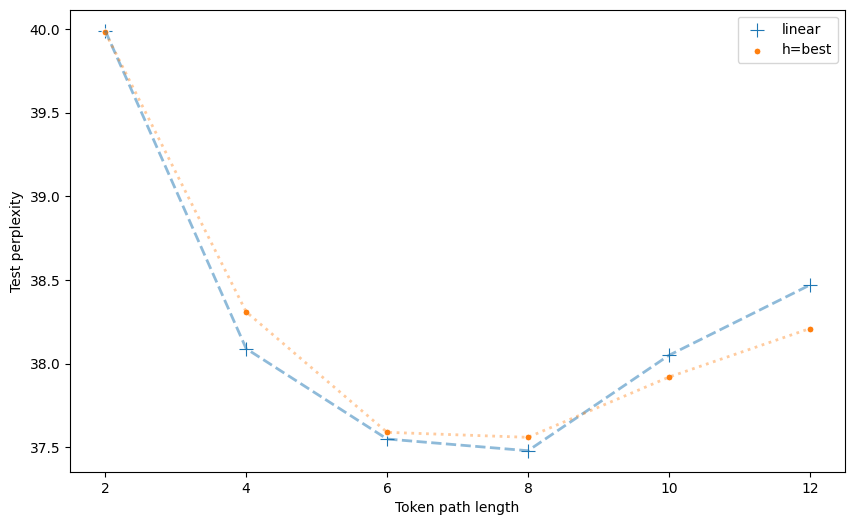}
            \label{fig:perf_linear_vs_best_h_combined}
        }
    }
     
    \subfloat [Combination of the WT2 and PTB]{ 
        \subfloat[Test perplexity]{
            \includegraphics[width=0.48\textwidth]{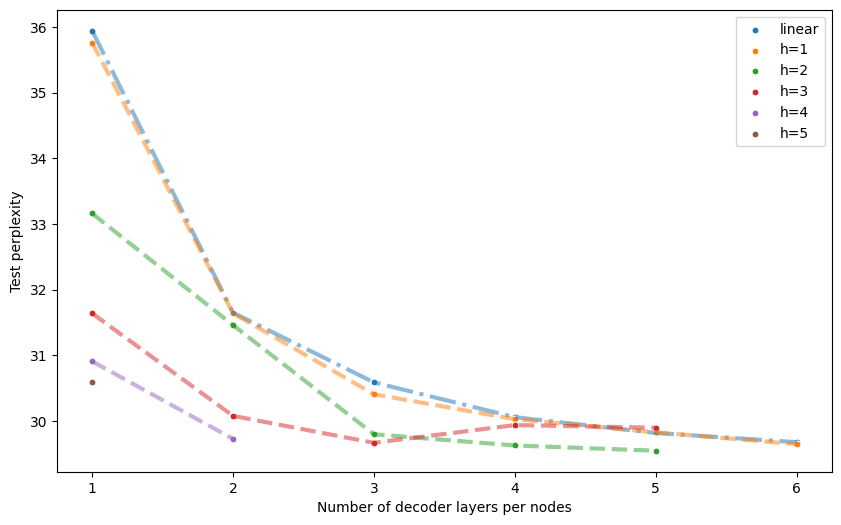}
            \label{fig:perf_height_combined}
        }
        \subfloat[Linear transformer vs best tree]{ 
            \includegraphics[width=0.48\textwidth]{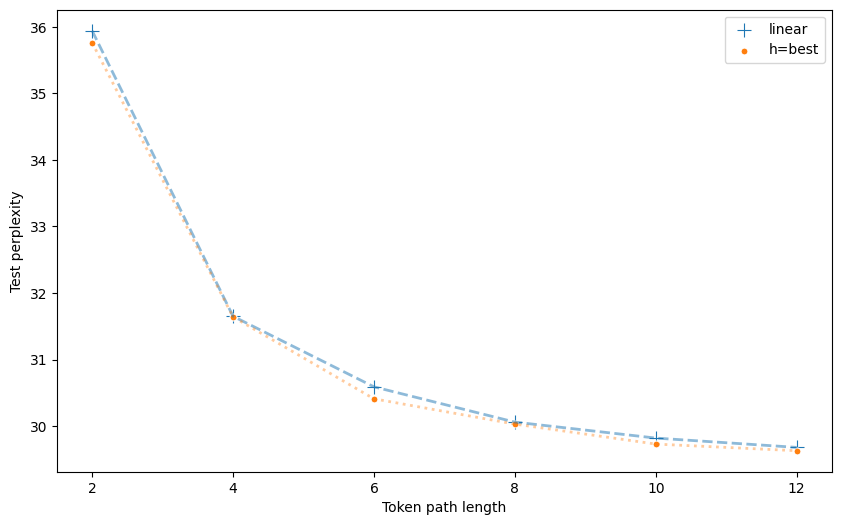}
            \label{fig:perf_linear_vs_best_h_combined}
        }
    }
    \caption{On the left, we compare the text perplexity for different height. On the right, we compare the test perplexity of a linear transformer with the best tree of the same path length, irrespective of height.  }
\end{center}
\end{figure}


Equally, increasing the height of the tree yields lower perplexity (reading the table horizontally) up to a certain threshold, beyond which the performance improvement stops. 
Although increasing the height increases the number of decoder layers a sequence has to go through, better performance was not always observed as it also increases the number of nodes and the number of paths a sequence can take and reduces the number of examples each path will see.

We can also group the results by the number of decoder layers a sequence has to go through, as seen in  Table \ref{tab:ppl-grouped}. For an equal number of layers, taller trees tend to outperform shorter ones. It suggests an improvement intrinsic to the tree structure.

\begin{table}[!h]
    \caption{Test perplexity for different architectures grouped by path length, the number of layers a token passes through. We also compare them with an equivalent linear transformer, the classic architecture without branching.}
    \label{tab:ppl-grouped}
    \centering
        \begin{tabular}{c@{\hspace{1cm}} c @{\hspace{0.5cm}}c@{\hspace{0.5cm}}c@{\hspace{0.5cm}}c*{7}{X} }
            \toprule
            Path length &  (h, dec) & Wikitext 02  &  PennTree  &  PennTree + WT2 \\
            
            \midrule
             
            4     & linear & 32.26  & 38.09 & 31.65 \\
                  & (1,2)  & 33.94  & 38.31 & 31.64 \\
                  & (3,1)  & 32.17  & 38.50 & 31.64\\
            \midrule
            
               & linear & 31.14  & 37.55 & 30.59 \\
               & (1,3)  & 30.99  & 37.83 & 30.41 \\
            6  & (2,2)  & 31.27  & 37.90 & 31.46 \\
               & (5,1)  & 31.26  & 37.59 & 30.59 \\
            \midrule
            
              & linear  & 30.80  &  37.48 & 30.06 \\
            8 & (1,4)   & 30.69  &  37.56 & 30.03 \\
              & (3, 2)  & 30.77  &  37.58 & 30.08 \\
            \midrule
            
               & linear & 30.56  & 38.05   &  29.82   \\
            10 & (1,5)  & 30.23  & 37.92   &  29.65 \\
               & (4, 2) & 30.70  & 38.01   &  29.73 \\
                 
            \midrule
            
                & linear & 30.67 & 38,47 &  29.68 \\
                & (1,6)  & 30.65 & 38.21 &  29.65 \\
            12  & (2, 4) & 30.68 & 38.38 &  29.63 \\
                & (3,3)  & 30.54 & 38.24 &  29.67  \\
            \midrule
            
            16  & linear & 30.39  & 39.31 &  tba   \\
                & (1,8)  & 30.67  & 39.10 &  29.83 \\
                & (3,4)  & 30.31  & 39.09 &  29.94  \\
                
            \bottomrule
        \end{tabular}
\end{table}




%
\subsection{ Random Selection. }
To ensure the selectors are actually learning a suitable routing behavior, i.e. sending down outputs to the better subtree, we compare our softmax selector with a random selector that chooses a child at random. 

Table \ref{tab:rand-v-selec} gathers results for a binary decoder of varying heights and number of layers per node and different datasets.
It shows that the softmax selector improves performance as expected, especially as the tree goes deeper. We can indeed see a 20-point difference for the tree of height 2 on the PennTreeBank dataset.

\begin{table}[!h]
\caption{Comparing softmax vs random selection.}
\label{tab:rand-v-selec}

\begin{tabularx}{\textwidth}{l @{\hspace{1.5cm}} l @{\hspace{1.5cm}} *{3}{X} }
\toprule
Dataset & Choice & \multicolumn{3}{c}{\thead{Tree Architectures with a 16 Path Length}} \\
\cmidrule(r){3-5}
& & (1,6) & (2,4) & (3,3) \\
\midrule
WikiText2 & random   & 35.39 & 40.21 & 46.74 \\
          & selector & 30.65 & 30.68 & 30.54 \\
\addlinespace
PennTreeBank & random   & 42.18 & 49.28 & 60.04 \\
             & selector & 38.21 & 30.38 & 38.24 \\
\addlinespace
WT2 + PTB    & random   & 33.51 & 42.04 & 49.11 \\
             & selector & 29.65 & 29.63 & 29.67 \\
\bottomrule
\end{tabularx}
\end{table}

\subsection{Influence of the selector size}
\begin{figure}[!ht]
\centering
\includegraphics[width=0.5\textwidth]{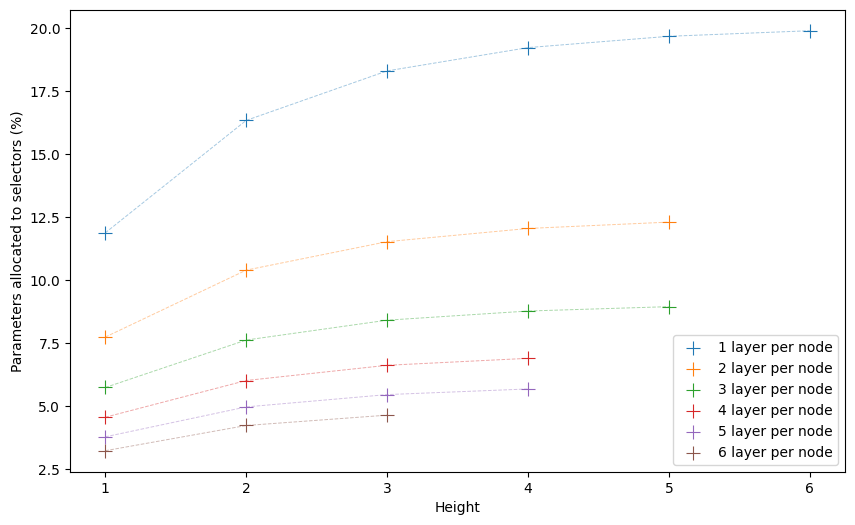}
\caption{Percentage of parameters allocated to the selection of the next child, with a hidden layer 8x the size of the input layer, for different numbers of layers per node and different heights.}
\label{fig:percent-selector-graph}
\end{figure}

We also studied the selector's influence on the model performance. We expect a bigger model to improve on the final test perplexity. The selector size plays a critical role in the model number of parameters as every time we increase the height, we increase the number of selectors as well. 
The question is of importance as the selector can account for a significant amount of parameters. In table \ref{tab:selec-percent}, we present the percentage of parameters of the model allocated to the selection for different architectures. It can lead to as much as 20\% dedicated to the next child node if we choose to have only 1 layer per node and could climb even higher if one decides to use a bigger branch selector. Thankfully, figure \ref{fig:percent-selector-graph} tells us that the percentage does not grow as fast as the tree. We will not allocate excessive resources to the selectors, as it would go against our premise of scaling our models while reducing inference time.  The results are seen in Table \ref{tab:selector-arch}.
 
\begin{table}[!h]
\caption{Percentage of parameters allocated to the selectors}
\label{tab:selec-percent}

\begin{tabularx}{\textwidth}{l *{12}{X} }
\toprule
	 & \multicolumn{6}{c}{\thead{Height of the tree}} \\
\cmidrule(){3-8}
Decoder& &  1       & 2       &     3    & 4  	     &    5     & 6  \\
\midrule
1   && 11.8	& 16.3	& 18.3	& 19.2	& 19.6	&19.9 \\
2   && 7.7	& 10.4	& 11.5	& 12.0	& 12.3	      \\
3   && 5.7	& 7.6	& 8.4	&  8.7	&  8.9	      \\
4   && 4.5	& 6.0	& 6.6	&  6.9		          \\
5   && 3.7	& 4.9	& 5.4	&  5.6		          \\
6   && 3.2	& 4.2	& 4.6	&		              \\
7   && 2.8	& 3.6	& 4.0	&		              \\
8   && 2.5	& 3.2	& 3.5	&		              \\
\bottomrule
\end{tabularx}
\end{table}

\begin{table}[!h]
\caption{Test perplexity for different selector architectures}
\label{tab:selector-arch}

\begin{tabularx}{\textwidth}{
  l
  @{\hspace{0.5cm}}
  l
  *{7}{>{\centering\arraybackslash}X} 
}
\toprule
Selector architecture &
\multicolumn{3}{c}{\thead{WikiText2}} &&
\multicolumn{2}{c}{\thead{PenTree  }} \\
\cmidrule(lr){2-4}
\cmidrule(lr){6-8}
                                     & (1,6) & (2,4) & (3,3) & & (1,6) & (2,4) & (3,3)\\
\midrule
Random                               & 35.39 & 40.21 & 46.74 & & 42.18 & 49.28 & 60.04  \\
Hidden layer 8x size of input layer  & 30.51 & 30.30 & 30.23 & & 38.21 & 38.38 & 38.24  \\
Hidden layer 16x size of input layer & 30.37 & 30.14 & 30.19 & & 38.40 & 38.22 & 38.12  \\
\bottomrule
\end{tabularx}
\end{table}

Our experiments demonstrate that scaling the network size does translate to improved test set perplexity.
Those improvements come at a cost (higher parameter count, heavier model, and slower inference) and the final decision to use a bigger network will always be a compromise and a choice left to the user.


%
\subsection{Influence of the branching factor }

The branching factor can be an interesting hyperparameter to vary as going from a binary to a ternary (or more) tree would increase the sparsity drastically. 

\begin{table}[!ht]
\caption{Test perplexity for multiple branching factors.}
\label{tab:branching}

\begin{tabularx}{\textwidth}{l@{\hspace{1cm}} l  *{7}{X} }
\toprule
 Dataset &  Factor  & \multicolumn{7}{c}{\thead{Tree Architecture\\(tree height, number of decoder layers per block) }} \\
\cmidrule(){3-9}
&                    & (1,1) & (1,2)  & (1,3)  & (1,4)  & (1,5)     & (2,1) & (2,2) \\
\midrule
 WikiText2	   & 2  & 37.09  & 33.94 & 30.99  & 30.69  &  30.23    & 33.45 & 31.27 \\
			& 3  & 39.96  & 32.32 & 30.94  & 30.64  &  31.07    & 37.25 & 31.20 \\
\addlinespace
PennTree        & 2  & 39.98  & 38.31 & 37.83  & 37.56  &  37.92    & 38.82 & 37.90\\
                & 3  & 40.15  & 38.44 & 37.60  & 37.49  &  38.00    & 39.05 & 37.57 \\
\addlinespace
\bottomrule
\end{tabularx}
\end{table}

Table \ref{tab:branching} shows that expanding our tree from binary to ternary can slightly improve performance for some architecture. Those benefits come at the expense of a much heavier model.




%
\section{ Future Work }
This work opens several promising avenues for future research. Firstly, we intend to explore diverse selection methods with distinct objectives. One direction of this investigation will focus on identifying the most potent model configuration, prioritizing the minimization of perplexity. Conversely, a parallel exploration will seek the most efficient architecture, acknowledging that even modest improvements beyond random selection (50\% perplexity) could represent a valuable trade-off in resource-constrained scenarios. Additionally, we aim to investigate the interplay between path optimality and model performance. This inquiry may reveal that selecting a good path, exhibiting demonstrably better performance than random selection outweighs the pursuit of an absolute best path. Furthermore, we will examine the potential trade-off between model length and optimality, evaluating whether longer, potentially less efficient models consistently outperform their shorter counterparts.
Furthermore, we propose exploring extreme tree configurations. This entails increasing both tree height and branching factor. By doing so, we aim to understand the emergence and impact of specialization within subtrees. 
We hypothesize that the architecture lends itself to efficient hardware parallelization. To test this hypothesis, we intend to conduct experiments where each node resides on a dedicated GPU, facilitating message-passing communication between decision points. These experiments will shed light on the practical scalability and hardware compatibility of the proposed architecture. Finally, while the current study focused on the architecture depicted in Figure \ref{fig:topo}\subref{fig:subfig-c}, further exploration of the alternative decoder-encoder combinations outlined in Figure \ref{fig:topo} is essential. 
\section{Conclusion}\label{sec13}
This study introduces TreeCoder, a novel class of tree-based transformers. The investigation primarily focused on a decoder-only architecture, demonstrating its promise by achieving performance on par with, or exceeding, its linear counterpart. Additionally, the decoder-only TreeCoder exhibits the advantageous properties of sparsity and inherent parallelization potential.
The study systematically explored the impact of architectural variations on perplexity, revealing that for equivalent layer counts, increasing the tree height provides greater benefit than expanding the number of layers within individual nodes.
Furthermore, the research demonstrates that all components of the tree-based architecture contribute meaningfully to the overall model behaviour, solidifying the utility and competitiveness of the proposed approach.

\backmatter

\begin{appendices}

\section{Hyperparameters}\label{secA1}

\begin{table}[!ht]
\centering
\caption{Hyperparameters used in this paper.}
\label{tab:hyperparam_summary}
\begin{tabular}{ll}
\hline
                       & Value         \\ \hline
Embedding size         & 1024          \\
Context length         & 128           \\
Learning rate          & 3e-4          \\
Batch size             & 16            \\
Dropout                & 0.1           \\ \hline
Optimizer              & AdamW         \\
$\beta_1$              & 0.9           \\
$\beta_2$              & 0.95          \\
$\epsilon$             & 1e-05         \\
Weight decay           & 0.01          \\
Gradient clipping      & 1             \\ \hline
Tokenizer              & SentencePiece \\
Model                  & BPE           \\
Vocab Size             & 8000          \\
Character Coverage     & 1             \\
Normalization rule     & Identity      \\
Split digit            & True          \\
Whitespace-only pieces & Allowed       \\
Byte fallback          & True          \\ \hline
Normalization          & RMSNorm       \\
$\epsilon$             & 1.e-5         \\ \hline
Activation function    & SwiGlu        \\
Seed                   & 42
                                  
\end{tabular}
\end{table}

\section{Results details}\label{secA2}

The table \ref{tab:ppl-h-dec} gathers the results of our experiments. "Layers" refers to the number of decoder layers in a node, and height is the height of the tree. The values are the test perplexity of models trained from scratch on the different datasets. The colors denote models with equal token path lengths. The token path lengths can be found in the sub-table \ref{fig:ppl-h-dec-d} .

\begin{table}[!ht]
    \begin{subfigure}{\textwidth}
        \begin{minipage}{0.48\textwidth}
            \begin{tabularx}{\textwidth}{c@{\hspace{0.5 cm}} *{5}{X} }
        	\toprule
        	Layers & \multicolumn{5}{c}{\thead{Height  }} \\
        	\cmidrule(){2-6}
        	& 1 & 2 & 3 & 4 & 5 \\
        	\midrule
        	1 & 37.09                       & 33.45                       & \colorbox{green!20}{32.17}  & 33.68                     & \colorbox{pink!20}{31.26}  \\
        	2 & \colorbox{green!20}{33.94}  & \colorbox{pink!20}{31.27}   & \colorbox{yellow!20}{30.77} & \colorbox{blue!20}{30.70} &                             \\
        	3 & \colorbox{pink!20}{30.99}   & 31.96                       & \colorbox{orange!20}{30.54} &                           &                             \\
        	4 & \colorbox{yellow!20}{30.69} & \colorbox{orange!20}{30.68} & \colorbox{gray!20}{30.31}   &                           &                              \\
        	5 & \colorbox{blue!20}{30.23}   & 31.44                       & 30.93                       &                           &                              \\
        	6 & \colorbox{orange!20}{30.65} & 30.79                       & 31.64                       &                           &                              \\
        	7 & 30.42                       &                             &                             &                           &                              \\
                8 & \colorbox{gray!20}{30.67}   &                             &                             &                           &                              \\
        	\bottomrule
            \end{tabularx}
            \caption{Wikitext2}
        \end{minipage} 
        \hfill
        \begin{minipage}{0.48\textwidth}
          
            \begin{tabularx}{\textwidth}{c@{\hspace{0.5 cm}} *{5}{X} }
        	\toprule
        	Layers & \multicolumn{5}{c}{\thead{Height  }} \\
        	\cmidrule(){2-6}
        	& 1 & 2 & 3 & 4 & 5 \\
        	\midrule
        	1 & 39.98                       & 38.82                       & \colorbox{green!20}{38.50}  &  38.25                       &  \colorbox{pink!20}{37.59}  \\
        	2 & \colorbox{green!20}{38.31}  & \colorbox{pink!20}{37.90}   & \colorbox{yellow!20}{37.58} &  \colorbox{blue!20}{38.01}   &                             \\
        	3 & \colorbox{pink!20}{37.83}   & 37.97                       & \colorbox{orange!20}{38.24} &                              &                             \\
        	4 & \colorbox{yellow!20}{37.56} & \colorbox{orange!20}{38.38} & \colorbox{gray!20}{39.09}   &                              &                             \\
        	5 & \colorbox{blue!20}{37.92}   & 38.62                       &                             &                              &                             \\
        	6 & \colorbox{orange!20}{38.21} &                             &                             &                              &                             \\
        	7 & 38.91                       &                             &                             &                              &                             \\
                8 & \colorbox{gray!20}{39.10}   &                             &                             &                              &                             \\
        	\bottomrule
            \end{tabularx}
            \caption{PennTreeBank}
        \end{minipage}
    \end{subfigure}
    \par\bigskip

    \begin{subfigure}{\textwidth}
        \begin{minipage}{0.48\textwidth}
            \begin{tabularx}{\textwidth}{c@{\hspace{0.5 cm}} *{5}{X} }
        	\toprule
        	Layers & \multicolumn{5}{c}{\thead{Height  }} \\
        	\cmidrule(){2-6}
        	& 1 & 2 & 3 & 4 & 5 \\
        	\midrule
        	1 & 35.76                       & 33.16                       & \colorbox{green!20}{31.64}  & 30.91                     & \colorbox{pink!20}{30.59}   \\
        	2 & \colorbox{green!20}{31.64}  & \colorbox{pink!20}{31.46}   & \colorbox{yellow!20}{30.08} & \colorbox{blue!20}{29.73} &                             \\
        	3 & \colorbox{pink!20}{30.41}   & 29.8                        & \colorbox{orange!20}{29.67} &                           &                             \\
        	4 & \colorbox{yellow!20}{30.03} & \colorbox{orange!20}{29.63} & \colorbox{gray!20}{29.94}   &                           &                             \\
        	5 & \colorbox{blue!20}{29.83}   & 29.55                       & 29.90                       &                           &                             \\
        	6 & \colorbox{orange!20}{29.65} &                             &                             &                           &                             \\
        	7 & 29.52                       &                             &                             &                           &                             \\
                8 & \colorbox{gray!20}{29.83}   &                             &                             &                           &                             \\
        	\bottomrule
            \end{tabularx}
            \caption{WT2 + PennTreeBank }
            \medskip \medskip
        \end{minipage} 
        \hfill
        \begin{minipage}{0.48\textwidth}
            \begin{tabularx}{\textwidth}{c@{\hspace{0.5 cm}} *{5}{X} }
        	\toprule
        	Layers & \multicolumn{5}{c}{\thead{Height  }} \\
        	\cmidrule(){2-6}
        	& 1 & 2 & 3 & 4 & 5 \\
        	\midrule
        	1 &	2	                     & 3	                      & \colorbox{green!20} 4	 & 5	                   & \colorbox{pink!20}6      \\
        	2 & \colorbox{green!20} 4	 & \colorbox{pink!20}6        & \colorbox{yellow!20}8	 & \colorbox{blue!20}{10}  & \colorbox{orange!20}{12} \\
        	3 &	\colorbox{pink!20}6	     & 9	                      & \colorbox{orange!20}{12} & 15	                   & 18                       \\
        	4 &	\colorbox{yellow!20}8	 & \colorbox{orange!20}{12}   & \colorbox{gray!20}{16}	 & 20	                   & 24                        \\
        	5 &	\colorbox{blue!20}{10}	 & 15	                      & 20	                     & 25	                   & 30                         \\
        	6 &	\colorbox{orange!20}{12} & 18	                      & 24	                     & 30	                   & 36                          \\
        	7 &	14	                     & 21	                      & 28	                     & 35	                   & 42                          \\
                8 & \colorbox{gray!20}{16}   & 24                         & 32                       & 40                      & 48                          \\
        	\bottomrule
            \end{tabularx}
            \caption{Total number of layers in a token's path  }
             \label{fig:ppl-h-dec-d}
        \end{minipage}
    \end{subfigure}
    \caption{Test perplexity for binary trees of different heights and number of decoder layers per node. Increasing either of these factors improves the perplexity up to certain thresholds, beyond which the performance improvement stops.}
    \label{tab:ppl-h-dec}
\end{table}

\end{appendices}

\printbibliography
\end{document}